\documentclass[11pt,twoside]{article}

\usepackage[left=2.5cm,right=2cm,top=2cm,bottom=2cm]{geometry}
\usepackage{fancyhdr} 
\usepackage{url} 
\usepackage[english]{babel}
\usepackage{amsmath}
\usepackage{graphicx}

\author{Zukang Liao, Stavros Petridis, Maja Pantic}
\title{Local Deep Neural Networks for Age and Gender Classification}

\begin{document}

\maketitle


\pagenumbering{roman}
\setcounter{page}{1}
\pagestyle{fancy}


\begin{abstract}

Local deep neural networks have been recently introduced for gender recognition. Although, they achieve very good performance they are very computationally expensive to train. In this work, we introduce a simplified version of local deep neural networks which significantly reduces the training time. Instead of using hundreds of patches per image, as suggested by the original method, we propose to use 9 overlapping patches per image which cover the entire face region. This results in a much reduced training time, since just 9 patches are extracted per image instead of hundreds, at the expense of a slightly reduced performance. We tested the proposed modified local deep neural networks approach on the LFW and Adience databases for the task of gender and age classification. For both tasks and both databases the performance is up to 1\% lower compared to the original version of the algorithm. We have also investigated which patches are more discriminative for age and gender classification. It turns out that the mouth and eyes regions are useful for age classification, whereas just the eye region is useful for gender classification.

\end{abstract}

\pagenumbering{arabic}
\setcounter{page}{1}
\fancyhead[LE,RO]{\slshape \rightmark}
\fancyhead[LO,RE]{\slshape \leftmark}

\section{Introduction and Related work}
Gender classification and age estimation can benefit a wide range of applications, e.g. visual surveillance, targeted advertising, human-computer interaction (HCI) systems, content-based searching etc \cite{applications}. In order to solve these two tasks accurately and efficiently, a modified version of the Local Deep Neural Networks (LDNN) \cite{LDNN} is proposed. The proposed method achieves similar results to most state-of-the-art methods while the amount of computation needed is largely reduced. We also investigate which face regions are important for gender/age-group classification. The conclusion drawn is that the eyes and mouth regions are the most informative ones for age classification, whereas just the eyes region is important for gender classification.

\subsection{Age and gender classification using CNNs}
A  convolutional neural network (CNN)  consisting of  three convolutional layers was used in \cite{CNN} to achieve 86.8$\pm$1.4\% and 50.7$\pm$5.1\% accuracy for gender and age-group classification, respectively, on the Adience database. Random patches of size 227 by 227 were cropped from the original and mirrored images in order to augment the training data and avoid overfitting. When testing, five 227-by-227 patches were generated from every single image. Four of them were aligned with the four corners of the image and one of them was aligned with the centre of the image. These five patches were then reflected horizontally, resulting in ten patches generated from every single image. The final prediction of the image was the average of these ten patches. The results of age classification on the Adience database were improved to 64.0$\pm$4.2\% using the VGG-16 CNN architecture pre-trained on ImageNet \cite{new_DEX}.

\subsection{Local deep neural networks (LDNNs)} \label{ssec:localDNN}

\begin{figure}[h]
\centering
\includegraphics[width =0.9\hsize]{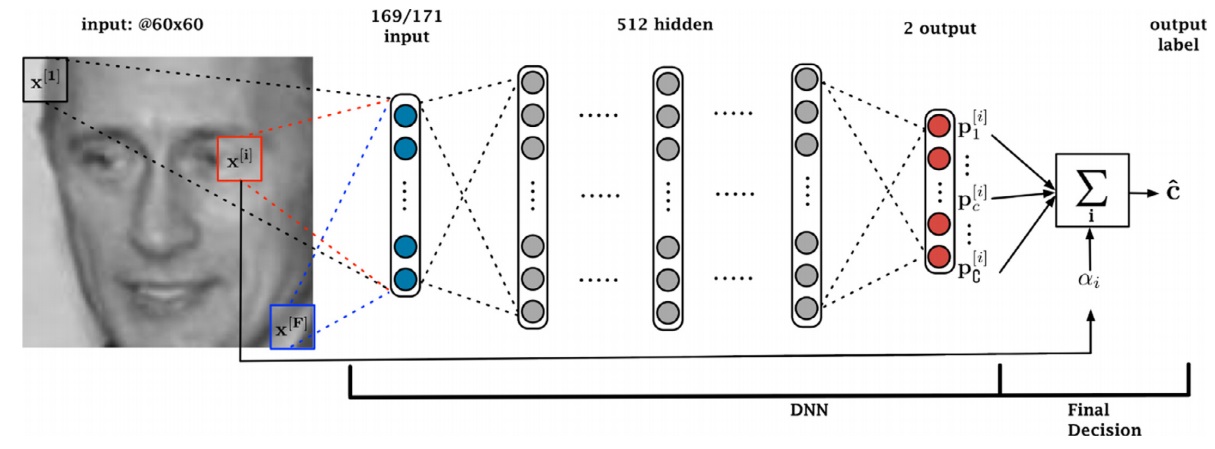}
\caption{Training a LDNN \cite{LDNN}.}
\label{fig:train_a_LDNN}
\end{figure}

LDNNs follow a different approach where small patches around important regions of images are extracted and then fed into deep neural networks to do gender/age classification \cite{LDNN}. The process is shown in Figure~\ref{fig:train_a_LDNN}. Initially, filters, e.g. edge or corner detectors, are used to find edges in images. Subsequently, patches are extracted around the detected edges. Finally, all patches are used to train neural networks. During testing, the predictions of all the patches obtained from one image are averaged as the final result for that image \cite{LDNN}. Classification rates of 96.25\% and 77.87\% in image and patch level, respectively, were achieved on the LFW database and 90.58\%  and 72.83\% in image and patch level, respectively, were achieved on the Gallagher's database using LDNN \cite{LDNN}. Using patches obtained in this way seldom leads to overfitting since most redundant information has been removed during filtering. Therefore, a simple feed forward neural network without dropout is used \cite{LDNN}.

\section{Databases}

\subsection{Labeled Faces in the Wild}
The Labeled Faces in the Wild (LFW) database contains 13,233 face photographs labeled with the name and gender of the person pictured. Images of faces were collected from the web with the only constraint that they were detected by the Viola-Jones face detector \cite{LFWTech}.
\bigskip
\begin{figure}[h]
\centering
\includegraphics[width =0.15\hsize]{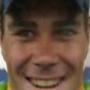}
\includegraphics[width =0.15\hsize]{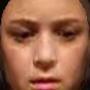}
\includegraphics[width =0.15\hsize]{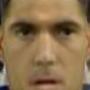}
\includegraphics[width =0.15\hsize]{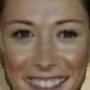}
\includegraphics[width =0.15\hsize]{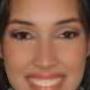}

\includegraphics[width =0.15\hsize]{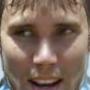}
\includegraphics[width =0.15\hsize]{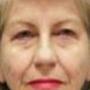}
\includegraphics[width =0.15\hsize]{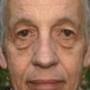}
\includegraphics[width =0.15\hsize]{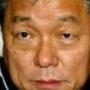}
\includegraphics[width =0.15\hsize]{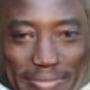}
\caption{Examples in the LFW 3D version.}
\label{fig:lfw_examples}
\end{figure}

There are four versions of LFW - the original version, funneled version \cite{lfw_funneled}, deep funneled version \cite{lfw_deepfunneled} and frontalised version (3D version). LFW is an imbalanced database including 10,256 images of men and 2,977 images of women from 5,749 subjects, 1,680 of which have two or more images \cite{LFWTech} \cite{LFWTechUpdate}. The 3D version is used in this work since the images are already cropped, aligned and frontalised properly as shown in Figure~\ref{fig:lfw_examples}.

\subsection{Adience database}
The Adience database contains 26,580 face photos from 2,284 individuals with gender and age labels of the person pictured. The images of faces were collected from the Flickr albums and released by their authors under the Creative Commons (CC) license. The images are completely unconstrained as they were taken under different variations in appearance, noise, pose, lighting etc \cite{Adience}.
\bigskip

\begin{figure}[h]
\centering
\includegraphics[width = 0.15\hsize]{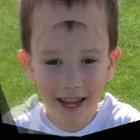}
\includegraphics[width = 0.15\hsize]{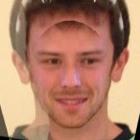}
\includegraphics[width = 0.15\hsize]{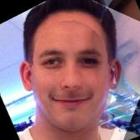}
\includegraphics[width = 0.15\hsize]{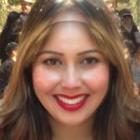}
\includegraphics[width = 0.15\hsize]{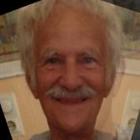}

\includegraphics[width = 0.15\hsize]{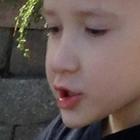}
\includegraphics[width = 0.15\hsize]{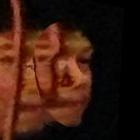}
\includegraphics[width = 0.15\hsize]{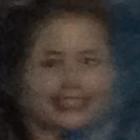}
\includegraphics[width = 0.15\hsize]{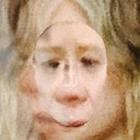}
\includegraphics[width = 0.15\hsize]{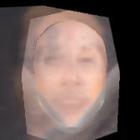}
\caption{Examples in the Adience database 3D version}
\label{fig:3d_ad_ex}
\end{figure}

There are three versions of the Adience database, including the original version, aligned version and frontalised version (3D version) with 26,580, 19,487 and 13,044 images respectively\cite{Adience}. The 3D version is used in this work since most images are already frontalised and aligned to the centre of the image. However, images in the Adience database 3D version may be extremely blurry or frontalised incorrectly as shown in Figure~\ref{fig:3d_ad_ex}. Additionally, people in the images could show emotions. Therefore, patches extracted from those images may  not always contain the same face region  which may  result in lower classification rates.

\bigskip

\subsubsection{Age groups merging}

There are eight age groups and another 20 different age labels in the Adience database as shown in Table \ref{AD_age_all}.  Using only the eight age groups is not feasible as no images for the $6^{th}$ group - (38, 43) exist in folds 1 or 2. Therefore, images that are labeled with one of the 20 age labels were merged to one of the eight age groups. In order to use all images labeled with age and to make the data more balanced, images were grouped as following:

\begin{itemize}
\item The $1^{st}$ age group: images labeled with '(0, 2)' or '2'
\item The $2^{nd}$ age group: images labeled with '(4, 6)' or '3'
\item The $3^{rd}$ age group: images labeled with '(8, 12)' or '13'
\item The $4^{th}$ age group: images labeled with '(15, 20)', '(8, 23)' or '22'
\item The $5^{th}$ age group: images labeled with '(25, 32)', '(27, 32)', '23', '29' or '34'
\item The $6^{th}$ age group: images labeled with '(38, 43)', '(38, 48)', '(32, 43)', '(38, 42)', '35', '36', '42' or '45'
\item The $7^{th}$ age group: images labeled with '(48, 53)', '55' or '56'
\item The $8^{th}$ age group: images labeled with '(60, 100)', '57' or '58'
\end{itemize}

\begin{table}[h]
\centering
\caption{All age labels in Adience database \cite{Adience}}
\label{AD_age_all}
\begin{tabular}{|c|c|c|c|c|c|}
\hline
          & Fold\_0 & Fold\_1 & Fold\_2 & Fold\_3 & Fold\_4 \\ \hline
(0, 2)    & 675     & 173     & 633     & 117     & 219     \\ \hline
(4, 6)    & 366     & 338     & 268     & 200     & 414     \\ \hline
(8, 12)   & 161     & 447     & 349     & 405     & 226     \\ \hline
(15, 20)  & 111     & 374     & 162     & 322     & 152     \\ \hline
(25, 32)  & 1,143   & 442     & 508     & 628     & 629     \\ \hline
(38, 43)  & 403     & 0       & 0       & 326     & 328     \\ \hline
(48, 53)  & 167     & 94      & 81      & 71      & 159     \\ \hline
(60, 100) & 94      & 98      & 129     & 80      & 182     \\ \hline
2         & 0       & 0       & 3       & 0       & 0       \\ \hline
3         & 12      & 0       & 2       & 3       & 0       \\ \hline
(8, 23)   & 0       & 0       & 0       & 1       & 0       \\ \hline
13        & 3       & 103     & 0       & 0       & 0       \\ \hline
22        & 0       & 77      & 2       & 7       & 0       \\ \hline
23        & 0       & 29      & 11      & 29      & 0       \\ \hline
(27, 32)  & 0       & 1       & 0       & 42      & 0       \\ \hline
29        & 0       & 0       & 6       & 0       & 0       \\ \hline
(32, 43)  & 0       & 300     & 208     & 0       & 0       \\ \hline
34        & 0       & 0       & 0       & 75      & 0       \\ \hline
35        & 21      & 80      & 8       & 77      & 28      \\ \hline
36        & 3       & 0       & 0       & 42      & 0       \\ \hline
(38, 42)  & 0       & 10      & 0       & 22      & 0       \\ \hline
(38, 48)  & 1       & 0       & 0       & 0       & 5       \\ \hline
42        & 0       & 0       & 0       & 1       & 0       \\ \hline
45        & 1       & 47      & 2       & 3       & 2       \\ \hline
55        & 3       & 44      & 3       & 3       & 8       \\ \hline
56        & 0       & 0       & 1       & 0       & 0       \\ \hline
57        & 0       & 0       & 2       & 4       & 10      \\ \hline
58        & 2       & 0       & 0       & 2       & 0       \\ \hline
\end{tabular}
\end{table}

We have defined this merging protocol so results  may not be directly comparable with other works that potentially use a different protocal. However, we were not able to find any publicly available protocol.

\subsubsection{One-off age classification rates}

Due to the similarity of people in adjacent age groups, images classified into adjacent age groups are also considered to be classified correctly and the corresponding result is called one-off classification rate \cite{CNN}.

\subsubsection{Three subsets of the Adience database 3D version}

Images labeled with gender are not necessarily labeled with age groups and vice-versa. This leads to three subsets of the Adience database 3D version. The $1^{st}$ subset consists of 12,194 images labeled with gender. The $2^{nd}$ subset consists of 12,991 images labeled with age. The $3^{rd}$ subset, where the experiments were conducted, consists of 12,141 images labeled with both gender and age as shown in Table \ref{3D_AD}.

\begin{table}[h]
\centering
\caption{Images labeled with both gender and age in the Adience database 3D version \cite{Adience}}
\label{3D_AD}
\begin{tabular}{|c|c|c|c|c|c|c|c|c|c|}
\hline
Group  &  $1^{st}$ &  $2^{nd}$ &  $3^{rd}$ &  $4^{th}$ &  $5^{th}$ &  $6^{th}$ &  $7^{th}$ &  $8^{th}$ & Total \\ \hline
Male   & 533    & 693    & 736     & 508     & 1,635   & 1,011    & 333      & 291       & 5,740 \\ \hline
Female & 494    & 910    & 952     & 699     & 1,867   & 875      & 296      & 308       & 6,401 \\ \hline
Total  & 1,027  & 1,603  & 1,688   & 1,207   & 3,502   & 1,886    & 629      & 599       & 12,141\\ \hline
\end{tabular}
\end{table}

\section{The proposed LDNN nine-patch method}

\begin{figure}[h]
\centering

\includegraphics[width = 0.59\hsize]{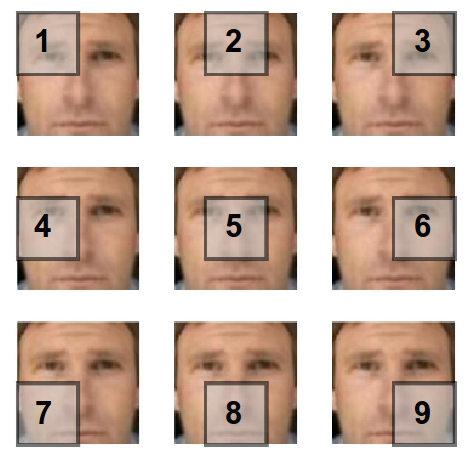}
\caption{The nine patches of the image}
\label{fig:ninePatchesOfTheImage}
\end{figure}

The LDNN approach (see Section \ref{ssec:localDNN}) generates hundreds of patches for every single image. As a consequence, it is computationally expensive to train a neural network when there are thousands of images. In order to reduce the computational cost, we propose the nine-patch method which generates only nine patches for every single image. Figure~\ref{fig:ninePatchesOfTheImage} shows an example of the nine patches of an image. The nine patches are indexed from left to right and then from top to bottom; The top left patch is indexed as the $1^{st}$ patch and the bottom right is indexed as the $9^{th}$ as shown in Figure~\ref{fig:ninePatchesOfTheImage}. The height and width of patches are set to half of the height and width of images. The overlapping ratio of adjacent patches is 50\%; the $2^{nd}$ patch overlaps 50\% with the $1^{st}$ and the $3^{rd}$ patch and the $4^{th}$ patch overlaps 50\% with the $1^{st}$ and the $7^{th}$ patch.

If images are frontalised and cropped properly, like images in the LFW database 3D version, the $1^{st}$ patch should certainly contain the left eye and the mouth should be contained in the $8^{th}$ patch.

\subsection{Pre-processing}

Initially, images are converted into grey-scale (from 0.0 - black to 1.0 - white). For images in the Adience database, the face is cropped using a fixed box - [20 20 100 100], which indicates that the coordinate of the top-left corner of the fixed box is [20 20] and the height and width of the box are both 100. For images in the LFW database, no box is needed since images are already cropped and aligned perfectly. Subsequently, images are resized to 60 by 60. Then the nine 30-by-30 patches are generated. For every single patch, pixel values were normalised to the standard Gaussian distribution with zero mean and unity variance.

\subsection{Methodology}
\begin{figure}[h]
\centering

\includegraphics[width = 1\hsize]{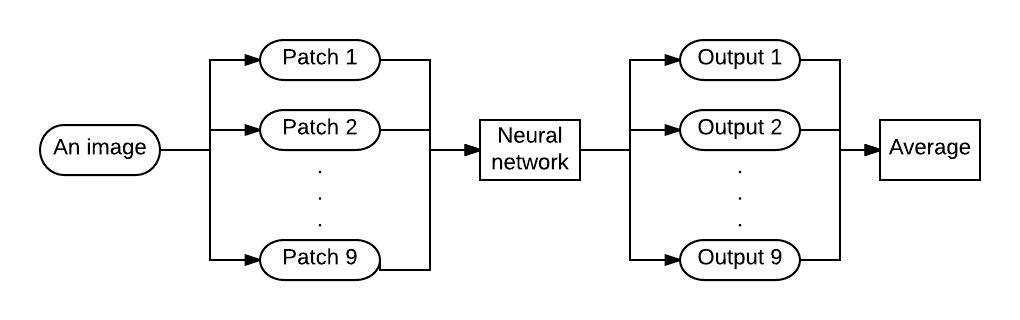}
\caption{The methodology of the nine-patch method}
\label{fig:methodology}
\end{figure}

During training, the nine patches are extracted as shown in Figure~\ref{fig:ninePatchesOfTheImage} and used to train a neural network. The testing procedure is shown in Figure~\ref{fig:methodology}. When testing, for every single image, the corresponding nine patches are classified by the trained neural network. Subsequently, the outputs or the posteriors of the nine patches are averaged. The averaged result is the final prediction of the image and the image will be classified to the class with the highest posterior.

\bigskip

\section{Experimental Results on the LFW database}

We have run a series of experiments on the LFW database, including optimisation of parameters and different combinations. In order to test the nine-patch method, five-cross-validation using the same five folds as \cite{LDNN} is used. Around $\frac{2}{3}$ of patches of men are randomly discarded in each fold to balance the data.

\subsection{Parameters of neural networks}
\label{sec:Parameters of neural networks}

We have run a series of experiments to identify a suitable set of training parameters. We have experimented with the number of hidden layers, number of hidden units per layer, dropout rates, activation functions (leaky ReLU), learning rate update policies etc. As a result, we have found that the following set of parameters leads to good performance:

\begin{itemize}
\item learning algorithm: SGD + Momentum
\item 0.8 and 0.5 retain dropout probability for input and hidden units, respectively
\item Initial learning rate: 3
\item Learning rate update rule: $lr \leftarrow lr*0.998$ from the first epoch to the last
\item Initial momentum: 0.5
\item Final momentum: 0.99
\item Momentum would increase from the first epoch until the $500^{th}$ epoch.
\item Number of hidden units: 512
\item Number of hidden layers: 2
\item Activation function: ReLU
\item Weights were initialised with He's method \cite{he2015delving}
\end{itemize}

\subsection{A replication of LDNN}
\begin{figure}[h]
\centering
\includegraphics[width = 1\hsize]{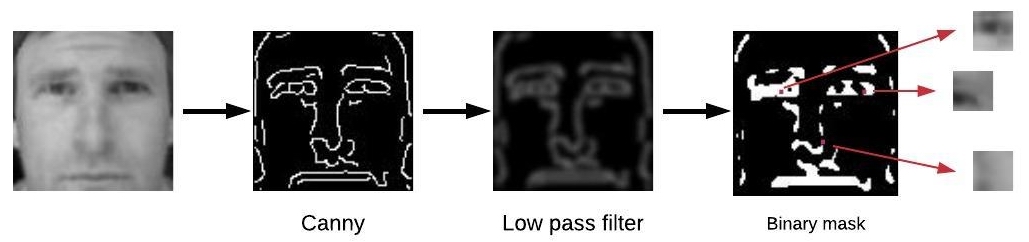}
\caption{Comparison between using the Sobel and Canny edge detector}
\label{fig:cannyandsobel}
\end{figure}

Firstly, the original LDNN method is replicated. After images have been cropped and resized to 60*60, a sobel/canny edge detector is used to obtain edges from the images. Subsequently, a low pass filter, e.g. a Gaussian filter, is used. Then a threshold is set so that strong edges, e.g. contours, would be preserved while noisy points and weak edges would be removed, resulting in the binary mask shown in Figure~\ref{fig:cannyandsobel}. Around every single white pixel in the binary mask, a patch with size 13*13 is generated.

There is a trade-off between the number of patches and the information left after pre-processing. When most useful edges are preserved, the number of patches is extremely large. For example, in the case shown in Figure~\ref{fig:cannyandsobel}, a nine-by-nine Gaussian kernel $N(\mu =0,\sigma =2)$ is used as the low pass filter and the threshold is set to 0.2. There are 425/485 patches on average for one image when the Sobel/Canny edge detector are used respectively as listed in Table \ref{table:cannyandsobel}, which leads to 425*2647 = 1,124,975 or 485*2647 = 1,283,795 patches respectively for a single fold of the LFW database. Due to the huge number of patches and the limited amount of memory, it is not feasible to train a neural network using all of the four training folds.
\bigskip

\begin{table}[h]
\centering
\caption{Comparison between our implementation of LDNNs and the original version presented in \cite{LDNN}. }
\label{table:cannyandsobel}
\begin{tabular}{|c|c|c|c|c|}
\hline
                            & Sobel   & Canny  & LDNN \cite{LDNN} & LDNN with location\cite{LDNN} \\ \hline
Number of patches per image & 425     & 485    & unknown & unknown \\ \hline
Patch level         & 63.27\% & 64.95\% & 74.50\% & 77.26\% \\ \hline
Image level         & 91.55\% & 93.68\% & 95.79\% & 96.25\% \\ \hline
\end{tabular}
\end{table}

Instead of using three or four folds, only one fold was used for training and one fold was used for testing, leaving three folds unused. This leads to classification rates of 91.55\%/93.68\% as shown in Table \ref{table:cannyandsobel}, which are lower than those reported in \cite{LDNN} $(\sim 96\%)$. However, this could well be the result of using a smaller training set that contains only one fold.

\begin{figure}[p]
\centering
\includegraphics[width = 0.19\hsize]{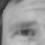}
\includegraphics[width = 0.19\hsize]{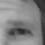}
\smallbreak

\includegraphics[width = 0.19\hsize]{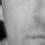}
\includegraphics[width = 0.19\hsize]{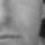}
\caption{Patch size 30}
\label{fig:patch_size_30}
\bigskip

\centering
\includegraphics[width = 0.12\hsize]{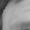}
\includegraphics[width = 0.12\hsize]{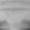}
\includegraphics[width = 0.12\hsize]{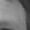}
\smallbreak

\includegraphics[width = 0.12\hsize]{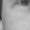}
\includegraphics[width = 0.12\hsize]{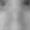}
\includegraphics[width = 0.12\hsize]{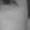}
\smallbreak

\includegraphics[width = 0.12\hsize]{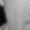}
\includegraphics[width = 0.12\hsize]{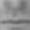}
\includegraphics[width = 0.12\hsize]{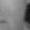}
\caption{Patch size 20}
\label{fig:patch_size_20}
\bigskip

\centering
\includegraphics[width = 0.09\hsize]{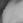}
\includegraphics[width = 0.09\hsize]{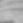}
\includegraphics[width = 0.09\hsize]{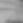}
\includegraphics[width = 0.09\hsize]{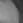}
\smallbreak

\includegraphics[width = 0.09\hsize]{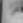}
\includegraphics[width = 0.09\hsize]{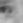}
\includegraphics[width = 0.09\hsize]{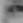}
\includegraphics[width = 0.09\hsize]{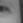}
\smallbreak

\includegraphics[width = 0.09\hsize]{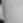}
\includegraphics[width = 0.09\hsize]{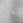}
\includegraphics[width = 0.09\hsize]{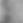}
\includegraphics[width = 0.09\hsize]{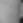}
\smallbreak

\includegraphics[width = 0.09\hsize]{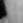}
\includegraphics[width = 0.09\hsize]{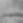}
\includegraphics[width = 0.09\hsize]{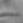}
\includegraphics[width = 0.09\hsize]{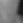}
\caption{Patch size 15}
\label{fig:patch_size_15}
\end{figure}

\subsection{Patch size and overlapping}

In order to identify the most suitable patch size, we have run a series of experiments with different patch sizes. The highest classification rate in image level without overlapping is obtained when patch size is 30 as shown in Table \ref{patch_size_512}. As shown in Figures~\ref{fig:patch_size_30} and \ref{fig:patch_size_15}, the reason that patch size 15 and 30 lead to over 94\% classification rates in image level could be that important facial regions, e.g. eyes, are not split into more than two patches. Especially when the patch size is 30, the $1^{st}$ patch of the first row would contain the entire left eye and the second would contain the right, which leads to the highest classification rate in image level without overlapping. Additionally, using patch size 30 requires the smallest amount of computation. Thus, patch size is set to 30.

\begin{table}[h]
\centering
\caption{Classification rates with different patch sizes and 512 hidden units for each hidden layer.}
\label{patch_size_512}
\begin{tabular}{|c|c|c|c|c|c|}
\hline
Patch Size  & 13*13    & 15*15    & 20*20    & 30*30   & Entire images \cite{LDNN}\\ \hline
Patch level & 73.12\%  & 75.13\%  & 78.50\%  & 87.56\% & 92.60\%\\ \hline
Image level & 93.366\% & 94.028\% & 92.852\% & 94.21\% & 92.60\%\\ \hline
\end{tabular}
\end{table}

Using 50\% overlapping ratio increases the classification rates by 1\% approximately while using 75\% overlapping ratio does not further improve the performance as shown in Table \ref{50_75_overlapping}. Therefore, the overlapping ratio is set to 50\%, which is also less computationally expensive.
\bigskip

\begin{table}[h]
\centering
\caption{Using overlapping ratios 0\%, 50\% and 75\% when patch size was 30}
\label{50_75_overlapping}
\begin{tabular}{|c|c|c|c|}
\hline
overlapping ratios & no overlapping & 50\%     & 75\%     \\ \hline
Patch level &87.56\% & 88.34\%  & 84.22\%  \\ \hline
Image level &94.21\% & 95.072\% & 94.25\%  \\ \hline
\end{tabular}
\end{table}

\subsection{Classification rate of individual patches}

For every single experiment using the nine patches, the classification rate of each patch is recorded. Results are shown in Table \ref{cr_Different9Patches}. It is consistent with all experiments that the highest classification rate comes from the patches that contain the eye.

\begin{figure}[h]
\centering
\includegraphics[width = 0.11\hsize]{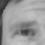}
\includegraphics[width = 0.11\hsize]{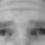}
\includegraphics[width = 0.11\hsize]{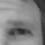}
\caption{The first, second and third patch from left to right}
\label{fig:patches_index1}

\includegraphics[width = 0.11\hsize]{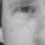}
\includegraphics[width = 0.11\hsize]{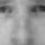}
\includegraphics[width = 0.11\hsize]{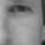}

\caption{The fourth, fifth and sixth patch from left to right}
\label{fig:patches_index2}

\includegraphics[width = 0.11\hsize]{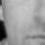}
\includegraphics[width = 0.11\hsize]{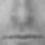}
\includegraphics[width = 0.11\hsize]{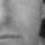}

\caption{The seventh, eighth and ninth patch from left to right}
\label{fig:patches_index3}
\end{figure}

\begin{table}[h]
\centering
\caption{Classification rates of patches}
\label{cr_Different9Patches}
\begin{tabular}{|c|c|c|c|c|c|c|c|c|c|}
\hline
                    & $1^{st}$       & $2^{nd}$       & $3^{rd}$       & $4^{th}$       & $5^{th}$       & $6^{th}$       & $7^{th}$       & $8^{th}$       & $9^{th}$       \\ \hline
Cr & 89.31\% & 87.75\% & 90.49\% & 90.11\% & 88.60\% & 90.84\% & 86.00\% & 86.01\% & 86.70\% \\ \hline
\end{tabular}
\end{table}

Compared with the last three patches that do not contain the eyes as shown in Figure~\ref{fig:patches_index1} to Figure~\ref{fig:patches_index3}, the classification rates of the first six patches are $2$ to $4\%$ higher.

\subsection{Five rows}
\begin{figure}[h]
\centering
\includegraphics[width = 0.225\hsize]{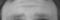}
\smallbreak
\includegraphics[width = 0.225\hsize]{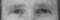}
\smallbreak
\includegraphics[width = 0.225\hsize]{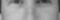}
\smallbreak
\includegraphics[width = 0.225\hsize]{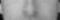}
\smallbreak
\includegraphics[width = 0.225\hsize]{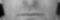}

\caption{Five 60*20 rows.}
\label{fig:60_20fiverows}
\end{figure}

In this experiment, five rows with 60-by-20 pixels are cropped for every single image as shown in Figure~\ref{fig:60_20fiverows}. Five neural networks are trained using one of the five rows separately. When testing, for every single image, the five rows are cropped and classified by the corresponding neural network. The final output of the image is the average of the outputs of the five rows cropped from the image.
\bigskip

Results are shown in Table \ref{5rows_cr}. As expected, the highest classification rate of a single row comes from the second row, which contains the two eyes as shown in Figure~\ref{fig:60_20fiverows}. Combining the five rows achieves 93.73\% classification rate, which is higher than any single row results. However, the performance is lower than the nine-patch method.

\begin{table}[h]
\centering
\caption{Classification rate of the five rows.}
\label{5rows_cr}
\begin{tabular}{|c|c|c|c|c|c|c|}
\hline
                    & The $1^{st}$ row   & The $2^{nd}$ row   & The $3^{rd}$ row   & The $4^{th}$ row   & The $5^{th}$ row & Combined \\ \hline
Accuracy & 89.73\% & 92.59\% & 90.58\% & 89.07\% & 90.24\% & 93.73\%  \\ \hline
\end{tabular}
\end{table}

\subsection{Using entire images}
Images are converted into greyscale and are resized to [32 32] and are mapped to the standard Gaussian distribution. The neural networks are then trained using entire images directly, which results in 92.72\% classification accuracy.

\subsection{Combinations}

\begin{figure}[h]
\centering
\includegraphics[width = 1\hsize]{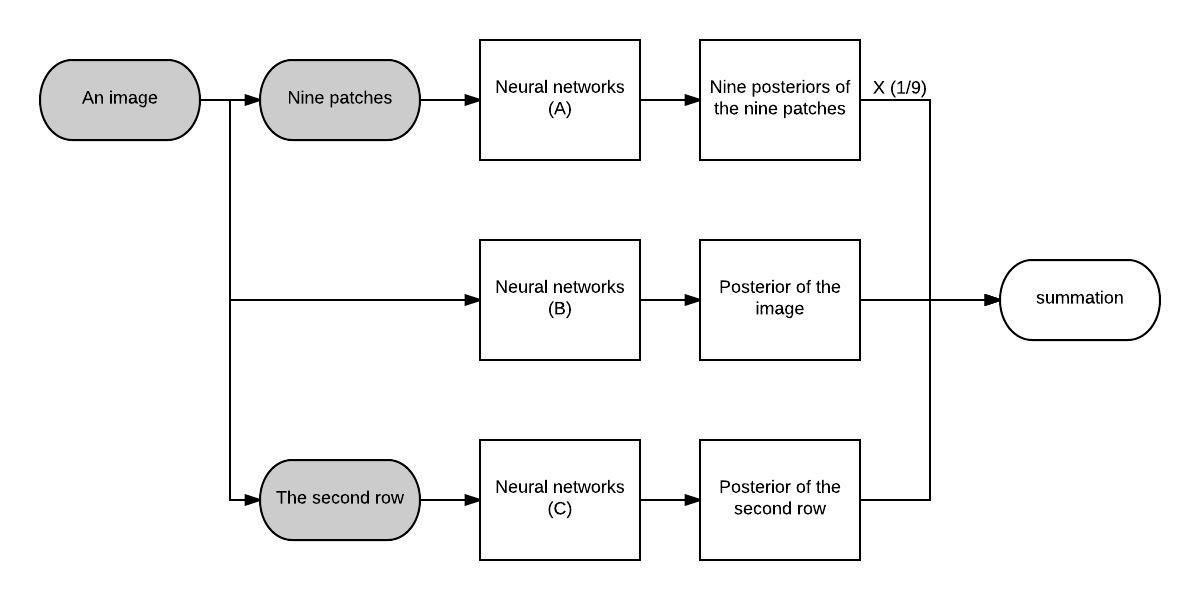}

\caption{Combining the nine-patch method with entire images and the second row}
\label{fig:2rowEntirePatches}
\end{figure}

In this section, we experiment with several combination of the nine patches, the entire images and the five rows. The best combination is shown in Figure~\ref{fig:2rowEntirePatches} and  results in a classification rate of 95.64\% as shown in Table \ref{comparisonCombinations}.
\bigskip

The best performance was achieved by combining the neural network (A) trained using the nine-patch method with the neural network (B) trained using entire images and the neural network (C) trained using the second row. When testing, for every single image, the nine patches are classified by the neural network (A) and the image itself is fed to the neural network (B) and the second row of the image is fed to the neural network (C). When combining these three sets of neural networks, the posteriors from the neural network (A) are multiplied by $\frac{1}{9}$, which means an entire image shares the same weight as the nine patches extracted from the image.
\bigskip

\begin{table}[h]
\centering
\caption{Comparison between the combination and using the nine-patch method only}
\label{comparisonCombinations}
\begin{tabular}{|c|c|c|c|}
\hline
         & Entire images & The nine-patch method & The combination \\ \hline
Accuracy & 92.72\%       & 95.072\%                     & 95.64\%                                                                    \\ \hline
\end{tabular}
\end{table}
\bigskip

To summarise, the highest classification rate that the nine-patch method can achieve on the LFW database is 95.072\% and combining the nine-patch method with entire images and the second row results in 95.64\%, both of which outperform the highest results that DNN, Deep Convolutional Neural Networks (DCNN), Gabor+PCA+SVM or BoostedLBP+SVM can achieve on the LFW database \cite{LDNN}. Additionally, compared with LDNN, the nine-patch method reduces the amount of computation largely while the classification rate is only around 0.5\% lower.

\section{Experimental results on the Adience database}

Initially, experiments for gender/age classification are conducted separately on the 3D version. Subsequently, the neural networks that are responsible for gender classification are used to assist age-group classification. Similarly, five cross validation using the same folds as \cite{CNN} is conducted to evaluate the performance.

\subsection{Gender classification}

We have trained neural networks using the nine-patch method and  entire images with the same parameters as in Section \ref{sec:Parameters of neural networks}. Results are shown in Table \ref{gendercrOnAdience}. The nine-patch method leads to a slight improvement ($1\%$ approximately) compared with training neural networks using entire images. Compared with the results on the LFW database, the gender classification rates achieved on the Adience database are approximately 17\% lower. The main reason is that images in the Adience database are not frontalised perfectly. As a consequence, patches do not always contain the same region of faces.

\begin{table}[h]
\centering
\caption{Results of gender classification}
\label{gendercrOnAdience}
\begin{tabular}{|c|c|c|}
\hline
         &  Using entire images & The nine-patch method \\ \hline
CR   & 77.79\%                                                                       & 78.63\%                                                                         \\ \hline
\end{tabular}
\end{table}

\begin{figure}[h]
\centering
\includegraphics[width = 0.11\hsize]{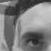}
\includegraphics[width = 0.11\hsize]{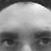}
\includegraphics[width = 0.11\hsize]{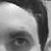}
\caption{The first, second and third patch from left to right}
\label{fig:patches_index_3DAdience1}

\includegraphics[width = 0.11\hsize]{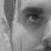}
\includegraphics[width = 0.11\hsize]{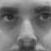}
\includegraphics[width = 0.11\hsize]{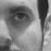}

\caption{The fourth, fifth and sixth patch from left to right}
\label{fig:patches_index_3DAdience2}

\includegraphics[width = 0.11\hsize]{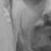}
\includegraphics[width = 0.11\hsize]{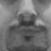}
\includegraphics[width = 0.11\hsize]{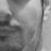}

\caption{The seventh, eighth and ninth patch from left to right}
\label{fig:patches_index_3DAdience3}
\end{figure}

The classification rates for each patch are shown in Table \ref{cr_DifferentPatches_Adience3D}. Unlike the LFW database, for the Adience database 3D version, higher classification rates only come from the second row, which consists of the $4^{th}$, the $5^{th}$ and the $6^{th}$ patches while the classification rates of the other six patches are 2\% - 4\% lower. The reason may be that the first three patches may not contain the eye or may be distorted seriously. For the second row (the $4^{th}$ to the $6^{th}$ patch), the patches contain the entire eye and a part of the nose as shown in Figures~\ref{fig:patches_index_3DAdience1} to \ref{fig:patches_index_3DAdience3}. Therefore, the classification rates of patches in the second row are the highest.

\begin{table}[h]
\centering
\centering
\caption{Gender classification rates of the nine patches on the 3D version}
\label{cr_DifferentPatches_Adience3D}
\begin{tabular}{|c|c|c|c|c|c|c|c|c|c|}
\hline
                    & $1^{st}$       & $2^{nd}$       & $3^{rd}$       & $4^{th}$       & $5^{th}$       & $6^{th}$       & $7^{th}$       & $8^{th}$       & $9^{th}$       \\ \hline
Cr & 70.82\% & 69.66\% & 71.55\% & 73.75\% & 72.15\% & 74.19\% & 70.98\% & 69.61\% & 71.42\% \\ \hline
\end{tabular}
\end{table}

\subsection{Age-group classification}
We also carry out age-group classification using  entire images and the nine-patch method. Results are shown in Table \ref{ResultsAge3D}. The age-group classification rate using the nine-patch method is 40.25\%. Compared with using entire images, the nine-patch method increases the classification rate by 0.75\% approximately and increases the one-off classification rate by 1\% approximately.
\bigskip

\begin{table}[h]
\centering
\caption{Results of age classification rates on the 3D version}
\label{ResultsAge3D}
\begin{tabular}{|c|c|c|}
\hline
         & Entire images (DNN) & The nine-patch method \\ \hline

Accuracy & 39.50\%                                                                      & 40.25\%                                                                        \\ \hline
One-off                                      & 76.18\%                                                                      & 77.05\%                                                                        \\ \hline
\end{tabular}
\bigskip

\centering
\caption{Age classification rates of the nine patches on the Adience database 3D version}
\label{cr_DifferentPatches_AdienceAligned_Age}
\begin{tabular}{|c|c|c|c|c|c|c|c|c|c|}
\hline
                    & $1^{st}$       & $2^{nd}$       & $3^{rd}$       & $4^{th}$       & $5^{th}$       & $6^{th}$       & $7^{th}$       & $8^{th}$       & $9^{th}$       \\ \hline
Cr & 35.07\% & 37.50\% & 35.64\% & 37.73\% & 39.52\% & 37.67\% & 33.52\% & 37.22\% & 34.53\% \\ \hline
\end{tabular}
\end{table}

The age-group classification rate of each patch is shown in Table \ref{cr_DifferentPatches_AdienceAligned_Age}. Similarly, the highest classification rates come from the patches in the second row that contains the eyes. In addition to the second row, the classification rates of the $2^{nd}$ patch and the $8^{th}$ patch are 2.5-3.5\% higher than those of the other patches. The reason may be that the $2^{nd}$ patch contains the inner corners of the two eyes and the $8^{th}$ patch contains the mouth.

\subsection{Age classification for men/women}

\begin{figure}[h]
\centering
\includegraphics[width = 1\hsize]{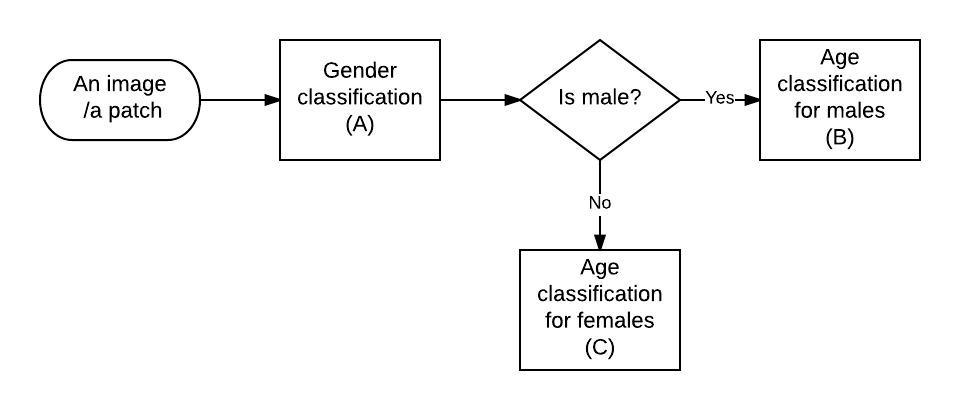}
\caption{Combination of age/gender classification}
\label{fig:CombineAgeGender}
\end{figure}

We conduct age-group classification for images of men and women separately using the nine-patch method and entire images. Results are shown in Table \ref{table:mnfAge}. In order to combine gender and age classification, the neural network (B) in Figure~\ref{fig:CombineAgeGender} is trained using 5,740 images of men only while the neural network (C) is trained using 6,410 images of women only.
\bigskip

\begin{table}[h]
\centering
\caption{Results of age-group classification for men and women}
\label{table:mnfAge}
\begin{tabular}{|c|c|c|c|c|}
\hline
                      & \multicolumn{2}{c|}{\begin{tabular}[c]{@{}c@{}}Neural network (B):\\ age-group classification for men\end{tabular}} & \multicolumn{2}{c|}{\begin{tabular}[c]{@{}c@{}}Neural network (C):\\ age-group classification for women\end{tabular}} \\ \hline
Classification rate   & Exact                                                     & One-off                                                    & Exact                                                     & One-off                                                     \\ \hline
Entire images         & 38.69\%                                                      & 77.63\%                                                    & 36.48\%                                                      & 74.83\%                                                     \\ \hline
The nine-patch method & 39.90\%                                                      & 80.32\%                                                    & 41.27\%                                                      & 77.14\%                                                     \\ \hline
\end{tabular}

\bigskip

\centering
\caption{Age classification rates of the nine patches of men}
\label{cr_DifferentPatches_AdienceAligned_AgeMen}
\begin{tabular}{|c|c|c|c|c|c|c|c|c|c|}
\hline
                    & $1^{st}$       & $2^{nd}$       & $3^{rd}$       & $4^{th}$       & $5^{th}$       & $6^{th}$       & $7^{th}$       & $8^{th}$       & $9^{th}$       \\ \hline
Cr & 30.84\% & 32.22\% & 31.07\% & 32.53\% & 34.42\% & 31.80\% & 32.01\% & 32.64\% & 31.08\% \\ \hline
\end{tabular}

\bigskip

\centering
\caption{Age classification rates of the nine patches of women}
\label{cr_DifferentPatches_AdienceAligned_AgeWomen}
\begin{tabular}{|c|c|c|c|c|c|c|c|c|c|}
\hline
                    & $1^{st}$       & $2^{nd}$       & $3^{rd}$       & $4^{th}$       & $5^{th}$       & $6^{th}$       & $7^{th}$       & $8^{th}$       & $9^{th}$       \\ \hline
Cr & 33.03\% & 34.37\% & 32.38\% & 34.49\% & 37.00\% & 34.37\% & 31.13\% & 34.01\% & 31.78\% \\ \hline
\end{tabular}
\end{table}

The classification rate of each patch is shown in Tables \ref{cr_DifferentPatches_AdienceAligned_AgeMen} and \ref{cr_DifferentPatches_AdienceAligned_AgeWomen}. For images of men, the classification rates of the $1^{st}$ and the $3^{rd}$ patch are almost the same as those of the $7^{th}$ and the $9^{th}$. However, for images of women, the classification rates of the $1^{st}$ and the $3^{rd}$ patch are about 2\% higher than those of the $7^{th}$ and the $9^{th}$. This indicates that the $1^{st}$ and the $3^{rd}$ patches, which contain the inner corners of the eyes, are more important to estimate women's age groups.
\bigskip

\begin{table}[h]
\centering
\caption{Testing age-group classification on other gender using the nine-patch method}
\label{table:wrongMFtesting}
\begin{tabular}{|c|c|c|}
\hline
        & \begin{tabular}[c]{@{}c@{}}Neural network (B) (Male):\\ tested on images of women \end{tabular} & \begin{tabular}[c]{@{}c@{}}Neural network (C) (Female):\\ tested on images of men \end{tabular} \\ \hline
Exact & 36.30\%                                                                                             & 36.15\%                                                                                               \\ \hline
One-off & 66.82\%                                                                                             & 73.07\%                                                                                               \\ \hline
\end{tabular}
\end{table}

\subsection{Combination of age/gender classification}

We run a series of experiments to combine gender and age-group classification. Results are shown in Table \ref{table:age/genderCombined} and the process is shown in Figure~\ref{fig:CombineAgeGender}. Three sets of neural networks already trained in the previous experiments are used. The neural network (A) is responsible for gender classification and the neural networks (B) and (C) are responsible for age-group classification for men and women, respectively. Initially, every single patch is classified by gender. If the patch is recognised as a patch of men, it would be fed and classified by the neural network (B). If the patch is recognised as a patch of women, it would be fed and classified by the neural network (C). Combining age/gender classification increases the classification rates by 1.5\% and increases the one-off classification rate by 1\% approximately.
\bigskip

\begin{table}[h]
\centering
\caption{Results of combining age/gender classification}
\label{table:age/genderCombined}
\begin{tabular}{|c|c|c|}
\hline
                    & The nine-patch method & The combination \\ \hline
Exact & 40.25\%                                                    & 41.82\%                                                                                 \\ \hline
One off             & 77.05\%                                                    & 77.98\%                                                                                                                  \\ \hline
\end{tabular}
\end{table}

If images/patches are classified incorrectly by the neural network (A) that is responsible for gender, they would be fed into the wrong neural network to do age-group classification. To investigate the effect on the performance of this issue, we fed images of men into the neural network (C) that is responsible for female age-group classification and vice-versa. Results are shown in Table \ref{table:wrongMFtesting}. In the former case, the classification rate decreases to 36.15\%/73.07\%(one-off) and in the latter case the age-group classification rate decreases to 36.30\%/66.82\%(one-off).
\bigskip

To summarise, using the nine-patch method, the gender and age-group classification rates on the Adience database 3D version are 78.63\% and 40.25\% respectively, which are approximately 1\% higher than using entire images. The combination of gender/age classification increases the age-group classification rates to 41.82\% and increases the one-off classification rate to 77.98\%. The results are similar to those established without using CNN \cite{CNN}.

\section{Conclusion}

We have presented a modified version of local deep neural networks which significantly reduces the training time and achieves classification rates which are around 0.5-1\% lower than the original version. Additionally, the proposed method outperforms other existing methods on the LFW database, e.g. Gabor+PCA+SVM \cite{Gabor+PCA+SVM} and BoostedLBP+SVM \cite{BoostedLBP+SVM}, and achieves similar results to other methods on the Adience database without using CNNs. We have also analysed the performance of different face regions and found that the eyes region is important for gender and age classification. In addition, the mouth region is also important for age estimation.

\bibliographystyle{plain}
\bibliography{ref}

\begin{thebibliography}{10}

\bibitem{Gabor+PCA+SVM}
Pablo Dago-Casas, Daniel Gonz{\'a}lez-Jim{\'e}nez, Long~Long Yu, and
  Jos{\'e}~Luis Alba-Castro.
\newblock Single-and cross-database benchmarks for gender classification under
  unconstrained settings.
\newblock In {\em Computer vision workshops (ICCV Workshops), 2011 IEEE
  international conference on}, pages 2152--2159. IEEE, 2011.

\bibitem{Adience}
Eran Eidinger, Roee Enbar, and Tal Hassner.
\newblock Age and gender estimation of unfiltered faces.
\newblock {\em IEEE Transactions on Information Forensics and Security},
  9(12):2170--2179, 2014.

\bibitem{he2015delving}
Kaiming He, Xiangyu Zhang, Shaoqing Ren, and Jian Sun.
\newblock Delving deep into rectifiers: Surpassing human-level performance on
  imagenet classification.
\newblock In {\em Proceedings of the IEEE international conference on computer
  vision}, pages 1026--1034, 2015.

\bibitem{lfw_funneled}
Gary~B. Huang, Vidit Jain, and Erik Learned-Miller.
\newblock Unsupervised joint alignment of complex images.
\newblock In {\em ICCV}, 2007.

\bibitem{lfw_deepfunneled}
Gary~B. Huang, Marwan Mattar, Honglak Lee, and Erik Learned-Miller.
\newblock Learning to align from scratch.
\newblock In {\em NIPS}, 2012.

\bibitem{LFWTech}
Gary~B. Huang, Manu Ramesh, Tamara Berg, and Erik Learned-Miller.
\newblock Labeled faces in the wild: A database for studying face recognition
  in unconstrained environments.
\newblock Technical Report 07-49, University of Massachusetts, Amherst, October
  2007.

\bibitem{LFWTechUpdate}
Gary B. Huang~Erik Learned-Miller.
\newblock Labeled faces in the wild: Updates and new reporting procedures.
\newblock Technical Report UM-CS-2014-003, University of Massachusetts,
  Amherst, May 2014.

\bibitem{CNN}
Gil Levi and Tal Hassner.
\newblock Age and gender classification using convolutional neural networks.
\newblock In {\em Proceedings of the IEEE Conference on Computer Vision and
  Pattern Recognition Workshops}, pages 34--42, 2015.

\bibitem{LDNN}
Jordi Mansanet, Alberto Albiol, and Roberto Paredes.
\newblock Local deep neural networks for gender recognition.
\newblock {\em Pattern Recognition Letters}, 70:80--86, 2016.

\bibitem{applications}
Choon~Boon Ng, Yong~Haur Tay, and Bok~Min Goi.
\newblock Vision-based human gender recognition: A survey.
\newblock {\em arXiv preprint arXiv:1204.1611}, 2012.

\bibitem{new_DEX}
Rasmus Rothe, Radu Timofte, and Luc Van~Gool.
\newblock Deep expectation of real and apparent age from a single image without
  facial landmarks.
\newblock {\em International Journal of Computer Vision}, pages 1--14, 2016.

\bibitem{BoostedLBP+SVM}
Caifeng Shan.
\newblock Gender classification on real-life faces.
\newblock In {\em International Conference on Advanced Concepts for Intelligent
  Vision Systems}, pages 323--331. Springer, 2010.

\end{thebibliography}
\end{document}